\documentclass[10pt,twocolumn,letterpaper]{article}

\usepackage{cvpr}
\usepackage{times}
\usepackage{epsfig}
\usepackage{graphicx}
\usepackage{amsmath}
\usepackage{amssymb}
\usepackage[percent]{overpic}
\usepackage{boldline}
\usepackage{multirow,enumitem}

\def\R{\mathbb R}

\usepackage[breaklinks=true,bookmarks=false]{hyperref}

\cvprfinalcopy % *** Uncomment this line for the final submission

\def\cvprPaperID{984} % *** Enter the CVPR Paper ID here

\ifcvprfinal\fi
\begin{document}

%%%%%%%%% TITLE
%\title{Efficient modalities fusion for video recognition with spatiotemporal semantic alignment }
\title{Improving the Performance of Unimodal Dynamic Hand-Gesture Recognition with Multimodal Training}

\author{Mahdi Abavisani\\%\thanks{Work started while at Microsoft}\\
Rutgers University\\
%Department of ECE, Piscataway, NJ 08854, USA\\
{\tt\small mahdi.abavisani@rutgers.edu}
\and
Hamid Reza Vaezi Joze\\
Microsoft\\
%One Microsoft way, Redmond, WA 98052, USA\\
{\tt\small hava@microsoft.com}
\and
Vishal M. Patel\\
Johns Hopkins University\\
%Department of ECE, Baltimore, MD  21218, USA\\
{\tt\small vpatel36@jhu.edu}
}

\maketitle
%\thispagestyle{empty}

%%%%%%%%% ABSTRACT
\begin{abstract}
We present an efficient approach for leveraging the knowledge from multiple modalities in training unimodal 3D convolutional neural networks (3D-CNNs) for the task of dynamic hand gesture recognition.   Instead of explicitly combining multimodal information, which is commonplace in many state-of-the-art methods, we propose a different framework in which we embed the knowledge of multiple modalities in individual networks so that each unimodal network can achieve an improved performance.  In particular, we dedicate separate networks per available modality and enforce them to collaborate and learn to develop networks with common semantics and better representations.  We introduce a ``spatiotemporal semantic alignment'' loss (SSA) to align the content of the features from different networks.  In addition, we regularize this loss with our proposed ``focal regularization parameter" to avoid negative knowledge transfer.   Experimental results show that our framework improves the test time recognition accuracy of unimodal networks, and provides the state-of-the-art performance on various dynamic hand gesture recognition datasets.  

% task tested on single modalities. Interestingly, we observe that  the improved unimodal networks contribute to an improved results in multimodal fusion as well. 
\end{abstract}

%%%%%%%%% BODY TEXT
\section{Introduction}
Recent advances in computer vision and pattern recognition have made hand gesture recognition an accessible and important interaction tool for different types of applications including human-computer interaction~\cite{rautaray2015vision}, sign language recognition~\cite{camgoz2017subunets}, and gaming and virtual reality control~\cite{lv2015touch}.   In particular, recent developments in deep 3-D convolutional neural networks (3D-CNNs) with video sequences have significantly improved the performance of dynamic hand gesture recognition \cite{neverova2014multi,molchanov2015hand,molchanov2016online}.

\begin{figure}
\begin{center}
\fbox{\begin{overpic}[width=0.45\textwidth,tics=3]{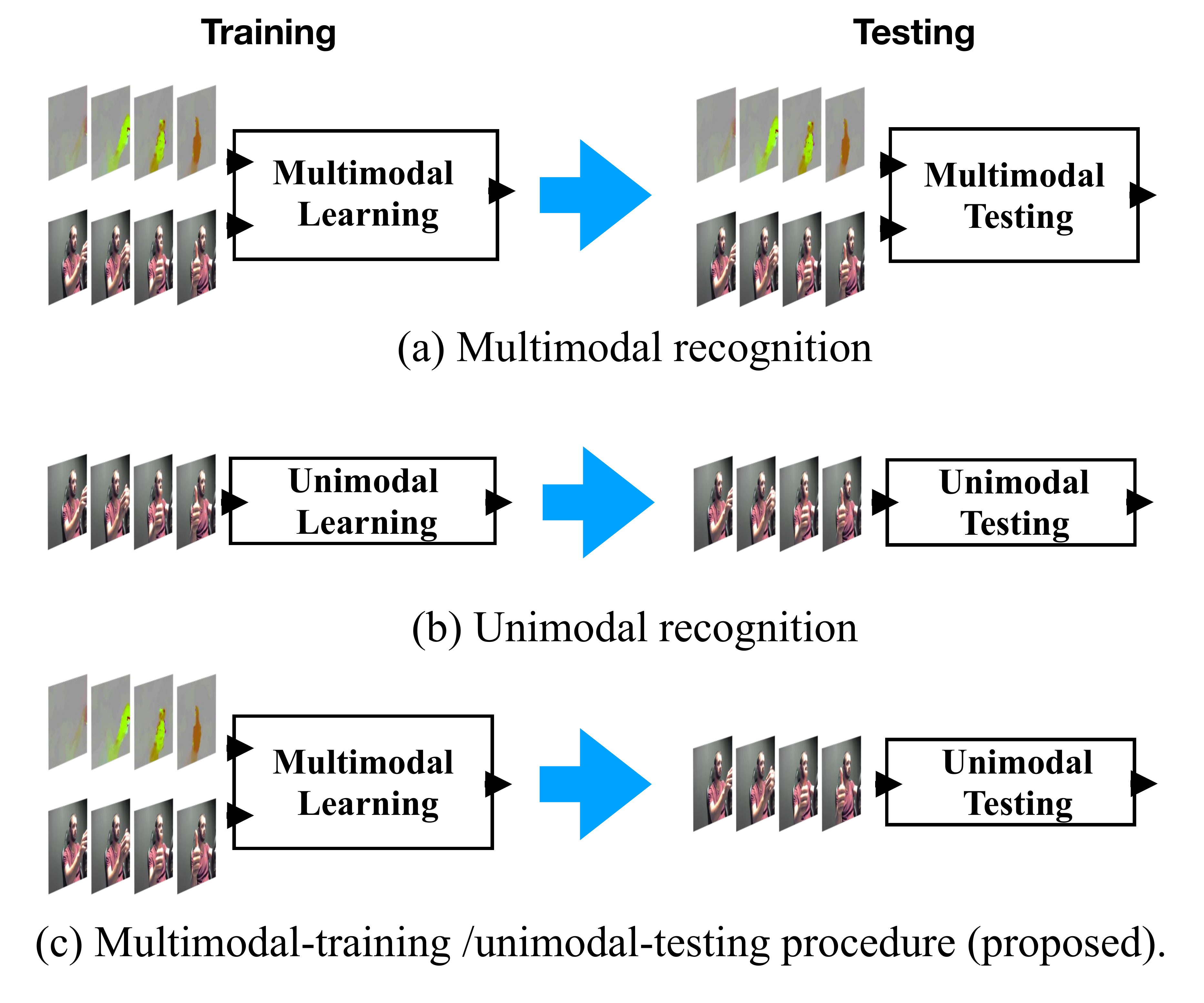}
% \put (29,69) {\small{(a) Multimodal recognition.}}
 %\put (30,39) {\small{(b) Unimodal recognition.}}
  %\put (0,4) {\small{(c) Multimodal-training/unimodal-testing procedure (proposed).}}
\end{overpic}
}
\end{center}
   \caption{Training and testing schemes of different types of recognition systems. (a) The system is trained and tested with multiple modalities. (b) The system is trained and tested with a single modality. (c) The system leverages the benefits of multimodal training but can be ran as a unimodal system during testing. }
\label{fig:definition}
\end{figure}

Most state-of-the-art hand gesture recognition methods exploit multiple sensors such as visible RGB cameras, depth camera or compute an extra modality like optical flow to improve their performances \cite{li2017lpsnet,zhang2017learning,wang2017large,miao2017multimodal}.  Multimodal recognition systems offer significant improvements to the accuracy of hand gesture recognition~\cite{molchanov2015multi}.  A multimodal recognition system is trained with multiple streams of data and classifies the multimodal observations during testing~\cite{ramachandram2017deep,babagholami2014multi} (Figure~\ref{fig:definition} (a)). On the other hand,  a unimodal recognition system is trained and tested using only a single modality data (Figure~\ref{fig:definition} (b)).   This paper introduces a third type of framework which leverages the knowledge from multimodal data during training and improves the performance of a unimodal system during testing.  Figure~\ref{fig:definition} (c) gives an overview of the  proposed framework. 

The proposed approach uses separate 3D-CNNs per each stream of modality for primarily training them to recognize the dynamic hand gestures based on their input modality streams.  The streams of modalities that are available in dynamic hand gesture recognition systems are often spatially and temporally aligned. For instance, the RGB and depth maps captured with motion sensing devices and the optical flow calculated from the RGB streams are usually aligned.  Hence, we encourage the individual modality networks to derive a common understanding for the spatiotemporal contents of different modalities.  We do this by sharing their knowledge throughout the learning process by minimizing the introduced \emph{spatiotemporal semantic alignment (SSA)} loss.    

We further improve the learning process by regularizing the SSA loss with an adaptive regularization parameter.  We call this regularization parameter, the \emph{focal regularization parameter}.  This parameter prevents the transfer of negative knowledge.  In other words, it makes sure that the knowledge is transferred from more accurate modality networks to less accurate networks and not the other way.  Once the networks are trained, during inference, each network has learned to recognize the hand gestures from its dedicated modality, but also has gained the knowledge transferred from the other modalities that assists in providing the better performance.  

In summary, this paper makes the following contributions. First, we propose a new framework for single modality networks in dynamic hand gesture recognition task to learn from multiple modalities.  This framework results in a \emph{Multimodal Training / Unimodal Testing (MTUT)} scheme.   Second, we introduce the \emph{SSA} loss to share the knowledge of single modality networks.   Third,  we develop the \emph{focal regularization parameter} for avoiding negative transfer.   In our experiments, we show that learning with our method improves the test time performance of unimodal networks.

%-------------------------------------------------------------------------
\section{Related Work}\label{sec:relatedwork}
\noindent \textbf{Dynamic Hand Gesture Recognition:}
Dynamic hand-gesture recognition methods can be categorized on the basis of the video analysis approaches they use.  Many hand-gesture methods have been developed based on extracting handcrafted features~\cite{wang2012robust,shen2012dynamic,tamrakar2012evaluation,ohn2014hand}.  These methods often derive properties such as appearance, motion cues or body-skeleton to perform gesture classification.    Recent advances in action recognition methods and the introduction  of various large video datasets have made it possible to efficiently classify unprocessed streams of visual data with spatiotemporal deep neural network architectures~\cite{carreira2017quo,tran2015learning,simonyan2014two}.    

Various 3D-CNN-based hand gesture recognition methods have been introduced in the literature.  A 3D-CNN-based method was introduced in ~\cite{molchanov2015hand} that integrates normalized depth and image gradient values to recognize dynamic hand gestures.  In~\cite{molchanov2015multi}, a 3D-CNN was proposed that fuses streams of data from multiple sensors including short-range radar, color and depth sensors for recognition.   A real-time method is proposed in ~\cite{molchanov2016online} that simultaneously detects and classifies gestures in videos.  Camgoz et al.~\cite{camgoz2016using} suggested a user-independent system based on the spatiotemporal encoding of 3D-CNNs.   Miao et al. proposed ResC3D~\cite{miao2017multimodal}, a 3D-CNN architecture that combines multimodal data and exploits an attention model. Furthermore, some CNN-based models also use recurrent  architectures to capture the temporal  information~\cite{zhang2017learning,cao2017egocentric,cui2017recurrent,zhu2017multimodal}.

 The main focus of this paper is to improve the performance of hand gesture recognition methods that are built upon 3D-CNNs.  As will be described later, we assume that our networks have 4-D feature maps that contain positional, temporal and channel dimensions.  
 
\noindent \textbf{Transfer Learning:}
In transfer learning, first, an agent is independently trained on a source task,  then another agent uses the knowledge of the source agent by repurposing the learned features or transferring them to improve its learning on a target task~\cite{pan2010survey,torrey2010transfer}.  This technique has been shown to be successful in many different types of applications  \cite{bengio2012deep,oquab2014learning,karpathy2014large,huang2013cross,yosinski2014transferable,Perera_CVPR19_1}.
While our method is closely related to transfer learning,  our learning agents (i.e. modality networks) are trained simultaneously, and the transfer occurs both ways among the networks.   Thus, it is better categorized as a multi-task learning framework~\cite{caruana1997multitask,oza2019deep}, where each network has three tasks of providing the knowledge to the other networks, receiving the knowledge from them, and finally classifying based on their dedicated input streams.

\noindent \textbf{Multimodal Fusion:}
In multimodal fusion, the model explicitly receives the data from multiple modalities and learns to fuse them~\cite{ngiam2011multimodal,abavisani2018multimodal,perera2018in2i}. The fusion can be achieved at feature level (i.e. early fusion), decision level (i.e. late fusion) or intermediately~\cite{ramachandram2017deep,abavisani2018deep}. Once the model is trained, during testing, it receives the data from multiple modalities for classification ~\cite{ramachandram2017deep,ngiam2011multimodal}.
While our method is related to multimodal fusion, it is not a fusion method.   We do not explicitly fuse the representations from different modalities. Instead, we improve the representation learning of our individual modality networks by leveraging the knowledge from different modalities.  During inference, we do not necessarily need multiple modalities but rather each individual modality network works independently to classify data.  

\begin{figure*}
\begin{center}
\fbox{\includegraphics[width=.9\linewidth]{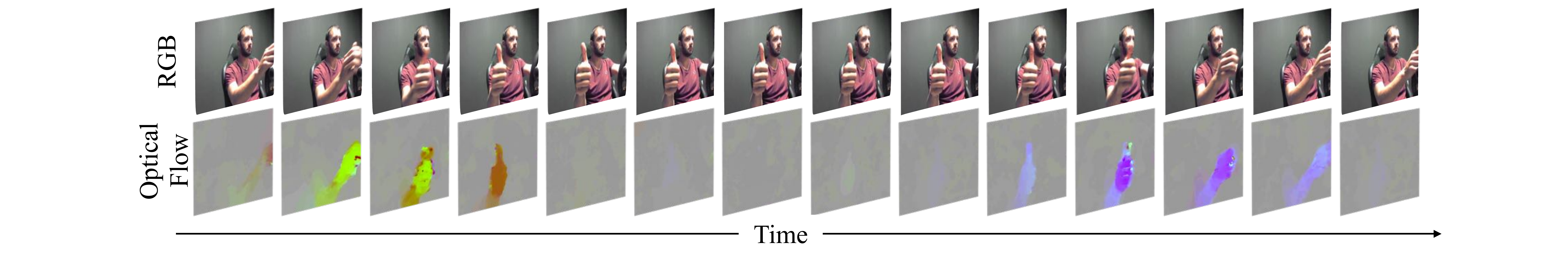}}
\end{center}
   \caption{An example of the RGB and optical flow streams from the NVGesture Dataset \cite{molchanov2016online}. As can be  seen, while for the stationary frames RGB provides better representation, optical flow provides better representation for the dynamic frames. }
\label{fig:2stream}
\end{figure*}

\section{Proposed Method}\label{sec:proposed}
 In our proposed model,  per each modality, one 3D-CNN is trained.  Assuming that the stream of data is available in $M$ modalities, we have $M$ classifier networks with similar architectures that classify based on their corresponding input.    During training, while each network is primarily trained with the data from its corresponding modality, we aim to improve the learning process by transferring the knowledge among the networks of different modalities.  The transferred knowledge works as an extra supervision in addition to the class labels.

We share the knowledge of networks by aligning the semantics of the deep representations they provide for the inputs.  We do this by selecting an in-depth layer in the networks and enforcing them to share a common correlation across the in-depth layers of all the modality networks. This is done by minimizing the distance between their correlation matrices in the training stage.  In addition, we regularize this loss by an adaptive parameter which ensures that the loss is serving as a one-way gate that only transfers the knowledge from more accurate modality networks to those with less accuracy, and not the other way.

\subsection{Spatiotemporal Semantic Alignment}
In an ideal case, all the $M$ classifier modality networks of our model should have the same understanding for an input video.  Even though they are coming in different modalities, their inputs are representing the same phenomena. In addition, since we assume that different modalities of the input videos are aligned over the time and spatial positions, in an ideal case the networks are expected to have the same understanding and share semantics for spatial positions and frames of the input videos across the different modalities. However, in practice, some spatiotemporal features may be better captured in one modality as compared to some other modalities. For instance, in the stream of visible RGB and optical flow frames shown in Figure~\ref{fig:2stream}, it can be observed that for static frames the RGB modality provides better information, while for dynamic frames optical flow has less noisy information. This results in different semantic understanding across the individual modality networks.

Thus, it is desirable to design a collaborative framework that encourages the networks to learn a common understanding across different modalities for the same input scene.  This way, if in a training iteration one of the networks cannot learn a proper representation for a certain region or time in its feature maps, it can use the knowledge from the other networks to improve its representations.  An iterative occurrence of this event during the training process leads the networks to develop better representations in a collaborative manner.

Let $\mathbf{F}_m , \mathbf{F}_n \in \R^{W\times H \times T \times C}$ be the in-depth feature maps of two networks corresponding to the $m$th modality and $n$th modality, where $W,H,T$ and $C$ denote width, heights, the number of frames and channels of the feature maps, respectively.   An in-depth feature map should contain high-level content representations (semantics) \cite{gatys2016image}. The element $\mathbf{f}^m_{i,j,t} \in \R^{C}$ in $\mathbf{F}_m$ represents the content for a certain block of time and spatial position.   It is reasonable to expect the network $m$ to develop correlated elements in $\mathbf{F}_m$ for spatiotemporal blocks with similar contents and semantics in the input.  Thus, in an ideal case, the correlated elements in $\mathbf{F}_m$ should have correlated counterpart elements in $\mathbf{F}_n$.

The correlations between all the elements of $\mathbf{F}_m$ is expressed by its correlation matrix defined as follows
\begin{equation}\label{eq:corr}
   \mathbf{corr}(\mathbf{F}_m) = \mathbf{\hat F}_m\mathbf{\hat F}_m^T  \in \R^{d \times d},
\end{equation}
where $\mathbf{\hat F}_m \in \R^{d \times C}$ contains the normalized elements of $\mathbf{F}_m$ in its rows, and $d =WHT$ is the number of elements in $\mathbf{F}^m$. The element $\mathbf{f}^m_{i,j,t}$ is normalized as  $\mathbf{\hat f}^m_{i,j,t} = {\mathbf{\tilde f}^m_{i,j,t}}/{\| \mathbf{\tilde f}^m_{i,j,t}\|}$ where $\| \mathbf{\tilde f}^m_{i,j,t}\|$ is the magnitude of $\mathbf{\tilde f}^m_{i,j,t}$, and $\mathbf{\tilde f}^m_{i,j,t}$ is calculated by $\mathbf{\hat f}^m_{i,j,t} = \frac{\mathbf{f}^m_{i,j,t} - \mu_{i,j,t}}{\sigma_{i,j,t}}$, where $\mu_{i,j,t}$ and $\sigma_{i,j,t}$ are respectively the sample mean and variance of the element.  We encourage the networks of the $m$th and the $n$th modalities to share a common correlation matrix for the feature maps of $\mathbf{F}_m$ and $\mathbf{F}_n$ so that they can have similar understanding for the input video while being free to have different styles.   We do this by minimizing their \emph{spatiotemporal semantic alignment} loss defined as
\begin{equation}\label{eq:ssa}
   \ell^{m,n}_{SSA} =  \rho^{m,n}\| \mathbf{corr}(\mathbf{F}_{m}) - \mathbf{corr}(\mathbf{F}_{n}) \|^2_{F}, 
\end{equation}
 where $\rho^{m,n}$ is an adaptive regularization parameter defined in Section \ref{sec:focal}.
 
 The \emph{spatiotemporal semantic alignment} loss is closely related to the covariance matrix alignment of the source and target feature maps in domain adaptation methods \cite{sun2016deep,sun2017correlation}.  In addition, in some style transfer methods, the Gram matrices of feature maps are aligned \cite{gatys2015texture,gatys2016image}. Aligning the Gram matrices, as opposed to our approach, discards the positional information and aligns the styles.  In contrast, our method aligns the positional and temporal information and discards the style.
 
 \begin{figure}
\begin{center}
%\fbox{\includegraphics[width=\linewidth]{focal_loss.pdf}}
\includegraphics[width=\linewidth]{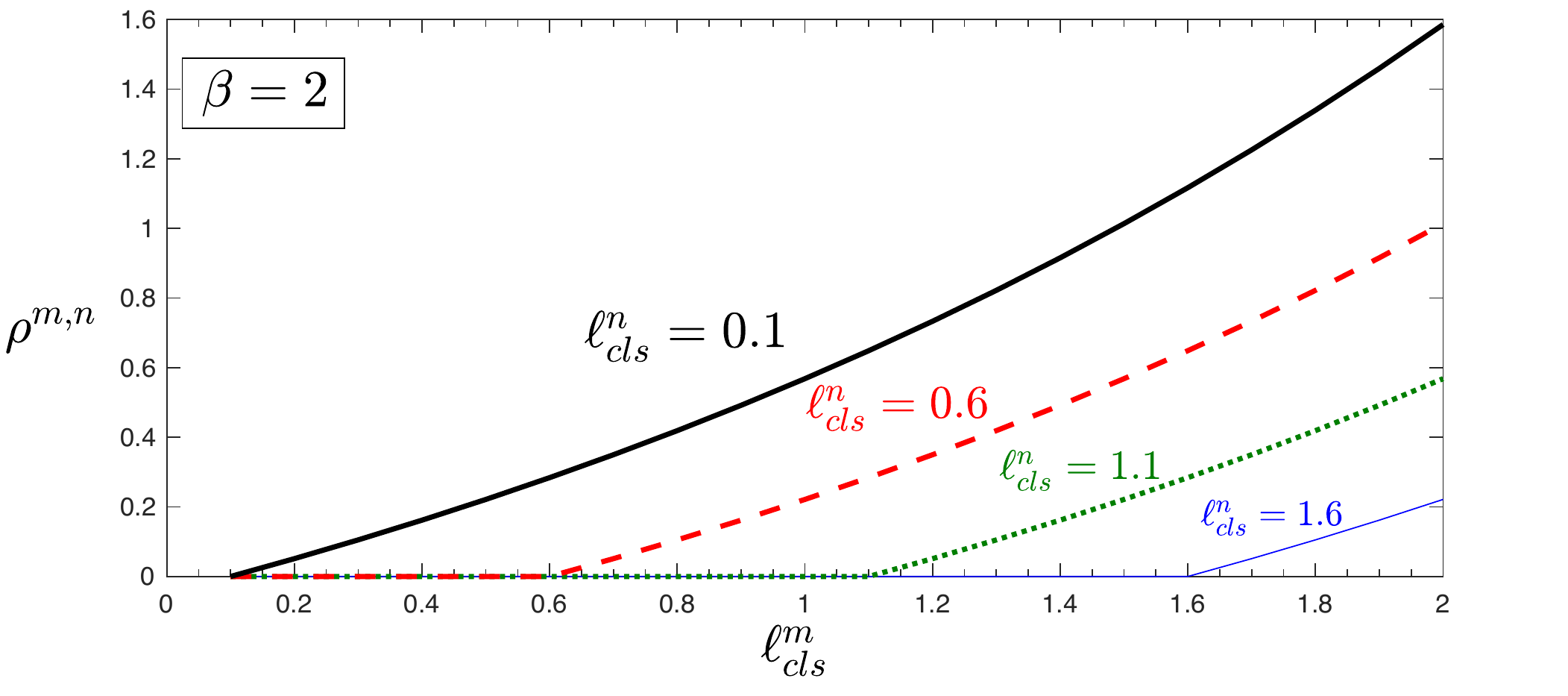}
\end{center}
   \caption{The value of focal regularization parameter ($\rho^{m,n}$) when  $\beta=2$ for different values of classification losses, $\ell^m_{cls}$ and $\ell^n_{cls}$.  Proportional to the classification performances of networks $m$ and $n$, this parameter scales the SSA loss to focus on transferring positive knowledge.}
\label{fig:focal}
\end{figure}

\subsection{Avoiding Negative Transfer}\label{sec:focal}
As discussed earlier, some modalities may provide weak features as compared to the others.  In addition, even the strong modalities may sometimes have corrupted or hard examples in their training set.    In these cases, aligning the spatiotemporal semantics of the representations from the other networks to the semantics of a week network may lead to a decrease in the performance. In such a case, a negative transfer has occurred. It is desirable to develop a method that produces positive knowledge transfer between the networks while avoiding negative transfer.   Such a method in our framework should enforce the networks to only mimic the semantics of more accurate networks in learning the representations for their hard examples.   To address this issue, we regularize our SSA loss with an adaptive regularization parameter termed as \emph{focal regularization parameter}.   This parameter is denoted as  $\rho^{m,n}$ in  equation~\eqref{eq:ssa}.

In order to measure the performance of the network modalities, we can use their classification loss values.  Assume $\ell^{m}_{cls}$ and $\ell^{n}_{cls}$ are the classification losses of the networks $m$ and $n$ that respectively correspond to the $m$th and $n$th modalities. In addition, let $\Delta \ell =\ell^{m}_{cls}  -   \ell^{n}_{cls}$ be their difference. A positive $\Delta \ell$ indicates that network $n$ works better than network $m$.  Hence,  in training the network $m$, for large positive values of $\Delta \ell$, we want large values for $\rho^{m,n}$ to enforce the network to mimic the representations of the network $n$.   As $\Delta \ell \rightarrow 0^+$, network $n$ becomes less an assist. Hence, we aim to have smaller $\rho^{m,n}$s to focus more on the classification task.   Finally, negative $\Delta \ell$ indicates that the network $n$ does not have better representations than the network $m$, and therefore $\rho^{m,n}$ should be zero to avoid the negative transfer.   To address these properties, we define the \emph{focal regularization parameter} as follows
\begin{equation}\label{eq:focalloss}
   \rho^{m,n} =  \mathbf{S}(e^{\beta \Delta\ell}   -1 ) =\left\{
\begin{array}{c l}	
     e^{\beta \Delta\ell}   -1   &  \Delta\ell>0\\
     0 &  \Delta\ell \leq 0
\end{array}\right.
\end{equation}
where $\beta$ is a  positive focusing parameter, and $\mathbf{S}(\cdot)$ is the thresholding function at zero.

%, and $e^{x}$ is the natural exponential function. 
Figure~\ref{fig:focal} visualizes values of $\rho^{m,n}$ for various $\ell^n_{cls}$s and $\ell^m_{cls} \in [0,2]$, when  $\beta=2$.   As can be seen, the parameter is dynamically scaled, where the scaling factor decays to zero as confidence in the classification performance of the current network modality increases (measured using $\Delta\ell$). This scaling factor can automatically down-weight the contribution of the shared knowledge if the performance of the modality network $n$ is degraded (measured by $\ell^n_{cls}$).

The focal regularization parameter $\rho^{m,n}$ is used as the regularization factor when aligning the correlation matrix of $\mathbf{F}^m$ in $m$th modality network to the correlation matrix of $\mathbf{F}^n$ in $n$th modality network.

\begin{figure}
\begin{center}
\fbox{\includegraphics[width=0.95\linewidth]{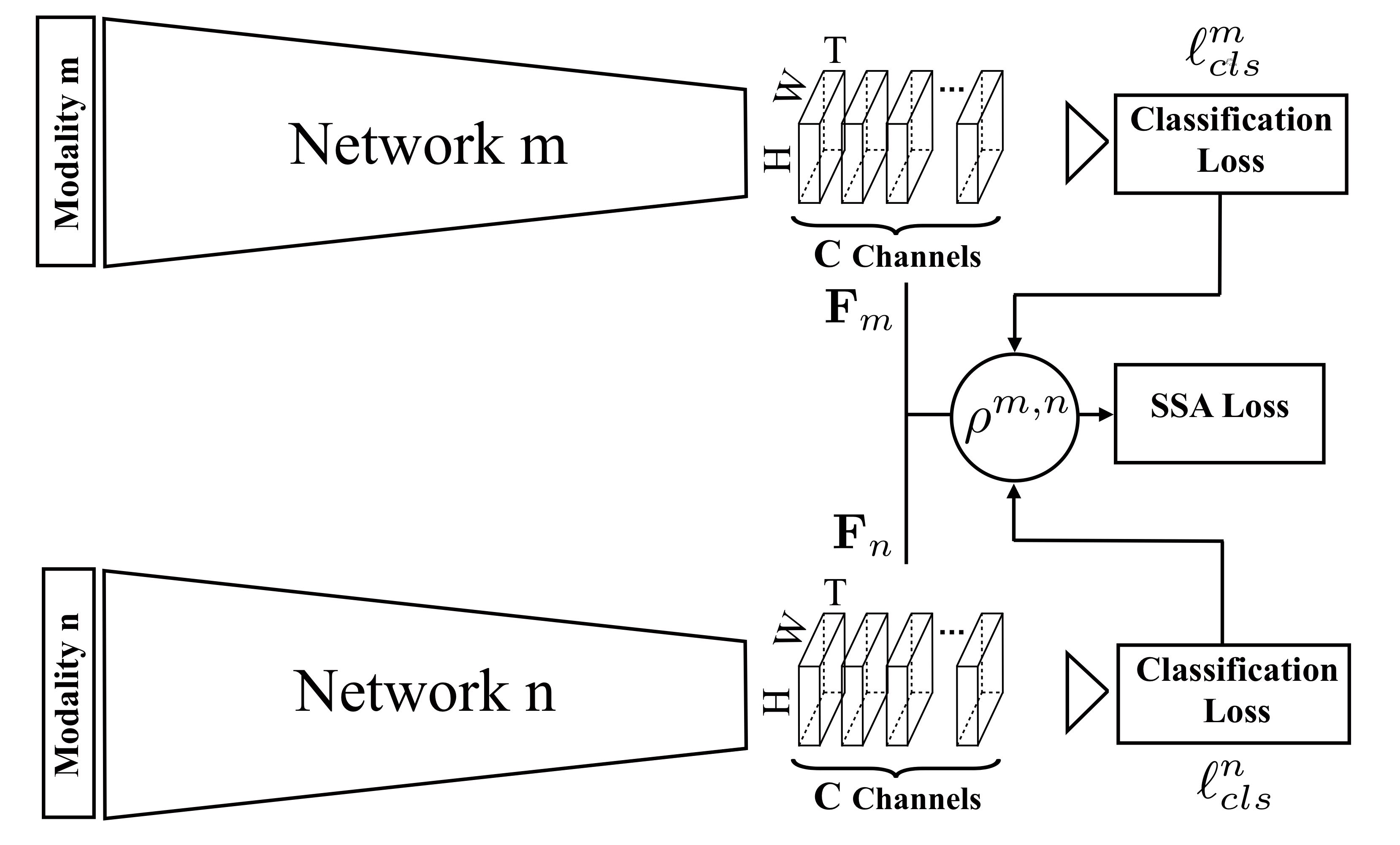}}
\end{center}
   \caption{Training network $m$ with the knowledge of network $n$.  Training network $m$, is primarily done with respect to its classifier loss ($\ell^m_{cls}$), but comparing with $\ell^n_{cls}$, $\rho^{m,n}$ determines if involving the \emph{SSA} loss is necessary, and if yes, it regularizes this loss with respect to the difference between the performances of two networks. Note that in the test time, both networks perform independently.}
\label{fig:diagram}
\end{figure}

\begin{figure*}
\begin{center}
\begin{overpic}[width=\textwidth,tics=3]{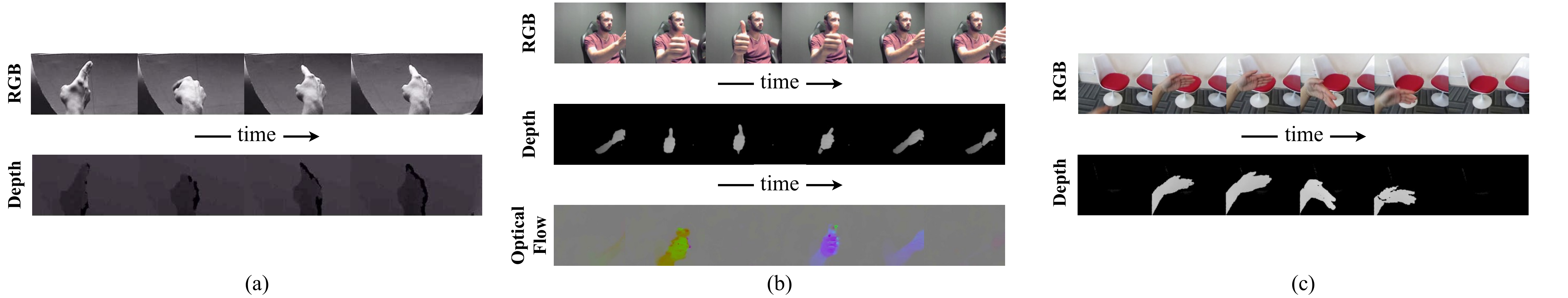}
\end{overpic}
\end{center}
\vspace{-4mm}
   \caption{Sample sequences from different modalities of used datasets. (a) VIVA hand gesture dataset~\cite{ohn2014hand}. (b) NVGesture dataset~\cite{molchanov2016online}. (c) EgoGesture~\cite{cao2017egocentric,zhang2018egogesture}.  As can be seen, the modalities in VIVA and EgoGesture datasets are well-aligned, while the depth map is not quite aligned with RGB and Optical flow maps in NVGesture.}
\vspace{-1mm}
\label{fig:datasets}
\end{figure*}

\subsection{Full Objective of the Modality Networks}
Combining the aforementioned objectives, our full objective for training the network corresponding to the $m$th modality in an $M$-modality task is as follows
\begin{equation}\label{eq:focalloss}
   \ell^{m} = \ell^m_{cls} + \lambda \sum_{n=1}^{M}\ell^{m,n}_{SSA}
\end{equation}
where $\lambda$ is a positive regularization parameter. Note that for $n=m$, $\rho^{m,n}=0$ and thus $\ell^{m,n}_{SSA} =0$. 

Figure \ref{fig:diagram} shows an overview of how the representations for the $n$th modality affects on learning the representation in the $m$th modality.  Since $\rho^{m,n}$ is differentiable, the training can be done in an end-to-end manner.    

Our model encourages the networks to improve their representation learning in the training stage.  During testing, each network performs separately.  Thus, once the networks are trained, one can use an  individual modality network to acquire efficient recognition.  However, it is worth mentioning that with our framework, applying a decision level modality fusion in the test stage is also possible. In fact, our experiments show that the proposed method not only improves the performance of unimodal networks, but it can also improve the fusion performance. 

%------------------------------------------------------------------------

\section{Experimental Results}\label{sec:result}
In this section, we evaluate our method against state-of-the-art dynamic hand gesture methods.  We conduct our experiments on three publicly available multimodal dynamic hand gesture datasets.  The following datasets are used in our experiments.
\begin{itemize}[noitemsep,topsep=0pt]
    \item \emph{VIVA hand gestures dataset}~\cite{ohn2014hand} is a multimodal dynamic hand gesture dataset specifically designed with difficult settings of cluttered background, volatile illumination, and frequent occlusion for studying natural human activities in real-world driving settings.   This dataset was captured using a Microsoft Kinect device, and contains 885 visible RGB and depth video sequences  (RGB-D) of 19 hand gesture classes, collected from 8 subjects.
    \item \emph{EgoGesture dataset}~\cite{cao2017egocentric,zhang2018egogesture} is a large multimodal hand gesture dataset collected for the task of egocentric gesture recognition.  This dataset contains 24,161 hand gesture clips of 83 classes of gestures, performed by 50 subjects.   Videos in this dataset include both static and dynamic gestures captured with an Intel RealSense SR300 device in RGB-D modalities across multiple indoor and outdoor scenes.
    \item \emph{NVGestures dataset}~\cite{molchanov2016online} has been captured with multiple sensors and from multiple viewpoints for studying human-computer interfaces. It contains 1532 dynamic hand gestures recorded from 20 subjects inside a car simulator with artificial lighting conditions.   This dataset includes 25 classes of hand gestures.  The gestures were recorded with SoftKinetic DS325 device as the RGB-D sensor and DUO-3D for the infrared streams.  In addition, the optical flow and infrared disparity map modalities can be calculated from the RGB and infrared streams, respectively.    We use RGB, depth and optical flow modalities in our experiments.  Note that IR streams in this dataset do not share the same view with RGB, depth and optical flow modalities. The optical flow is calculated using the method presented in~\cite{farneback2003two}. 
\end{itemize}

Figure~\ref{fig:datasets} (a), (b), and (c) show sample frames from  the different modalities of these datasets that are used in our experiments.  Note that the RGB and depth modalities are well-aligned in the VIVA and EgoGesture datasets, but are not completely aligned in the NVGestures dataset.

For all the datasets, we compare our method against two state-of-the-art action recognition networks, I3D~\cite{carreira2017quo} and C3D~\cite{tran2015learning}, as well as state-of-the-art dynamic hand gesture recognition methods that were reported on the used datasets. In the tables, we report the results of our method as~\emph{``Multimodal Training Unimodal Testing'' (MTUT)}.

\noindent \textbf{Implementation Details: } In the design of our method, we adopt the architecture of I3D network as the backbone network of our modality networks, and employ its suggested implementation details \cite{carreira2017quo}.  This network is an inflated version of Inception-V1~\cite{ioffe2015batch}, which contains several 3D convolutional layers followed with 3D max-pooling layers and inflated Inception-V1 submodules. The detailed architecture can be found in~\cite{carreira2017quo}.   We select the output of the last inflated Inception submodule,  ``$\texttt{Mixed\_5c}$'', as the in-depth feature map in our modality networks for applying the SSA loss~\eqref{eq:ssa}. In all the experiments $\lambda$ is set to $50\times 10^{-3}$, and  $\beta=2$. The threshold function in the focal regularization parameter is implemented by a ReLu layer.   For all the experiments with our method and I3D benchmarks, unless otherwise stated, we start with the publicly available ImageNet~\cite{deng2009imagenet} + Kinetics~\cite{kay2017kinetics} pre-trained networks.

%\footnote{The pre-trained models on ImageNet+Kinetics for RGB and optical flow streams are available at: \url{https://github.com/deepmind/kinetics-i3d}. For the networks dedicated to depth map stream, we also start with the RGB pre-trained model.}. 

We set the momentum to 0.9, and optimize the objective function with the standard SGD optimizer.   We start with the base learning rate of $10^{-2}$ with a $10\times$ reduction when the loss is saturated.  We use a batch size of 6 containing 64-frames snippets in the training stage. The models were implemented in Tensor-Flow 1.9~\cite{abadi2016tensorflow}. For our method, we start with a stage of pretraining with only applying the classification losses on the modality networks for $60$ epochs, and then continue training with the SSA loss for another $15$ epochs.

%Similar to~\cite{carreira2017quo} w
We employ the following  spacial and temporal data augmentations during the training stage. For special augmentation, videos are resized to have the smaller video size of 256 pixels, and then randomly cropped with a $224\times224$ patch.   In addition, the resulting video is randomly but consistently flipped horizontally.  For temporal augmentation,  64 consecutive frames are picked randomly from the videos.   Shorter videos are randomly padded with zero frames on both sides to obtain 64 frames.  During testing, we use $224\times224$ center crops, apply the models convolutionally over the full video, and average predictions.

Note that we follow the above mentioned implementation details identically for the experiments with both the I3D method~\cite{carreira2017quo}, and our method.  The only difference between the I3D method and our MTUT is in their learning objective.  In our case, it consists of the introduced constraints as well.

\iffalse
 \begin{figure}
\begin{center}
%\fbox{\includegraphics[width=\linewidth]{focal_loss.pdf}}
\includegraphics[width=.8\linewidth]{objfn.pdf}
\end{center}
   \caption{The objective function values of different methods vs iterations.}
\label{fig:objfn}
\end{figure}
\fi

\subsection{VIVA Hand Gestures Dataset}
%VIVA hand gesture~\cite{ohn2014hand} with only 885 video samples is relatively small for training 3D-CNNs with millions of parameters.  Thus,  knowledge transfer can play an essential role in providing desirable performance on this dataset.   
In this set of experiments, we compare our method on the VIVA dataset against a hand-crafted approach (HOG+HOG2~\cite{ohn2014hand}), a recurrent CNN-based method (CNN:LRN~\cite{molchanov2015hand}), a C3D~\cite{tran2015learning} model which were pre-trained on Sports-1M dataset~\cite{karpathy2014large} as well as the I3D method that currently holds the best results in action recognition~\cite{carreira2017quo}.     All the results are reported by averaging the classification accuracies over 8-fold cross-subject cross-validation.

Table~\ref{tbl:viva_unimodal} shows the performance of the dynamic hand gesture methods tested on the visible and depth modalities of the VIVA dataset.   As can be seen from this table, the I3D network performs significantly better than  \emph{HOG+HOG2} and \emph{CNN:LRN}. This is in part due to the knowledge that I3D contains from its pretraining on ImageNet and Kinematic datasets.    Nonetheless,  we observe that our method that shares the same architecture and settings with the I3D networks and only differs in the learning procedure has significantly improved the I3D method by a $3.08\%$ boost in the performance of RGB's network and $6.85\%$ improvement on the performance of the depth's network. This experiment shows that our method is able to integrate the complementary information between two different modalities to learn efficient representations that can improve their individual performances.
 \begin{table}[t]
\begin{center}
%\resizebox{\linewidth}{!}{%
\begin{tabular}{ l c c}
\hlineB{3}
%\cline{3-7}
& \multicolumn{2}{c}{\small{Testing modality}} \\
\cline{2-3} 
 Method   & RGB & Depth\\% & Fusion (RGB-D)\\%   & Avg.          $\pm$  std.      \\

\hline
 %\parbox[t]{2mm}{\multirow{6}{*}{\rotatebox[origin=c]{90}{C3D backbone}}}
HOG+HOG2~\cite{ohn2014hand}  &	52.3	& 	58.6\\%	& 	64.5		 \\
CNN:LRN~\cite{molchanov2015hand} &	57.0	& 	65.0\\%	& 	74.4		 \\
%CNN:LRN:HRN~\cite{molchanov2015hand} &	-	& 	-	& 	77.5		 \\
C3D~\cite{tran2015learning} &	71.26 	& 	68.32	\\%& 	77.4		 \\
I3D~\cite{carreira2017quo}       &	78.25	& 	74.46	\\%& 	83.10 \\
MTUT~(ours) &	\bf{81.33}	& 	\bf{81.31}	\\%& 	\bf{86.08}		 \\
\hlineB{3}
\end{tabular}
\vspace{2mm}
\caption{8-fold cross-subject average accuracies of different hand gesture methods on the VIVA hand gesture dataset~\cite{ohn2014hand}. The top performer is denoted by boldface. } \label{tbl:viva_unimodal}
\vspace{-2mm}
\end{center}
%\squeezeupbig\squeezeupbig\squeezeup
\end{table}

\subsection{EgoGesture Dataset}
We assess the performance of our method along with various hand gesture recognition methods published on the large-scale hand gesture dataset, EgoGesture~\cite{cao2017egocentric}.  Table~\ref{tbl:ego_unimodal} compares unimodal test accuracies of different hand gesture methods.  VGG16~\cite{simonyan2014very} is a frame-based recognition method,  and VGG16+LSTM~\cite{donahue2015long} combines this method with a recurrent architecture to leverage the temporal information as well.  As can be seen, the 3D-CNN-based methods, C3D, C3D+LSTM+RSTMM~\cite{cao2017egocentric}, and I3D, outperform the  VGG16-based methods.  However, among the 3D-CNN-based methods, our method outperforms the top performers in the RGB domain by $2.15\%$ and in the Depth domain by  $1.09\%$.

 \begin{table}[t]
\begin{center}
%\resizebox{\linewidth}{!}{%
\begin{tabular}{ l c c}
\hlineB{3}
%\cline{3-7}
& \multicolumn{2}{c}{\small{Testing modality}} \\
\cline{2-3} 
 Method   & RGB & Depth\\% & Fusion (RGB-D)\\%   & Avg.          $\pm$  std.      \\

\hline
VGG16~\cite{simonyan2014very} &	62.5	& 	62.3\\% 	& 	66.5		 \\
VGG16 + LSTM~\cite{donahue2015long} &	74.7	& 	77.7\\% 	& 	81.4		 \\
C3D~\cite{tran2015learning} &	86.4	& 	88.1\\% 	& 	89.7		 \\
C3D+LSTM+RSTTM~\cite{cao2017egocentric} &	89.3	& 	90.6	\\% & 	92.2		 \\
I3D~\cite{carreira2017quo} &	90.33	& 	89.47\\% 	& 	92.78		 \\
MTUT~(ours) &	\bf{92.48}	& 	\bf{91.96}\\% 	& 	\bf{93.87}		 \\
\hlineB{3}
\end{tabular}
\vspace{1mm}
\caption{Accuracies of different hand gesture methods on the EgoGesture dataset~\cite{cao2017egocentric}. The top performer is denoted by boldface.} \label{tbl:ego_unimodal}
\vspace{-1mm}
\end{center}
%\squeezeupbig\squeezeupbig\squeezeup
\end{table}

 \begin{figure}[htp!]
	\begin{center}
		%\fbox{\includegraphics[width=\linewidth]{focal_loss.pdf}}
		\includegraphics[width=.95\linewidth]{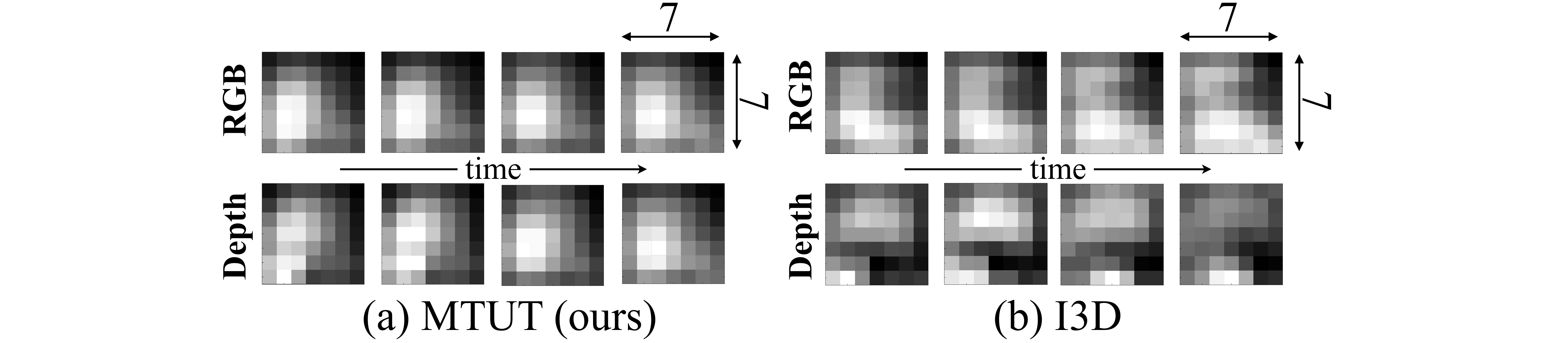}
	\end{center}
	\caption{Visualization of the feature maps corresponding to the layer ``$\texttt{Mixed\_5c}$'' in different networks for a sample input from EgoGesture dataset.   These figures show the sequence of average feature maps (over 1024 channels) in  (a) the RGB  and depth networks trained with the I3D method.  (b) the RGB  and depth networks trained with our method. Intensity displays the magnitude.}
	\label{fig:featuremaps}
\end{figure}

In Figure~\ref{fig:featuremaps}, we visualize a set of feature maps from the RGB and depth networks trained with the I3D and our method.  We feed a given input from the EgoGesture dataset to different networks and calculate the average of feature maps over the channels in the layer ``$\texttt{Mixed\_5c}$''.   We display the resulting sequence in four $7 \times 7$ blocks.  Here the temporal dimension is four and the spatial content is $7 \times 7$.  Layer ``$\texttt{Mixed\_5c}$'' is the layer in the I3D architecture in which we apply the SSA loss to.   We observe that the networks trained with our model have learned to detect similar structures for the given input (Figure~\ref{fig:featuremaps} (a)). On the other hand, the networks trained with the I3D model are not bounded to develop similar structures. Thus, even though the input of the two modalities represent the same content, the feature maps may detect different  structures (Figure~\ref{fig:featuremaps} (b)).
%\vspace{-2.5mm}

\begin{figure}
\begin{center}
\begin{overpic}[scale=0.15,tics=3]{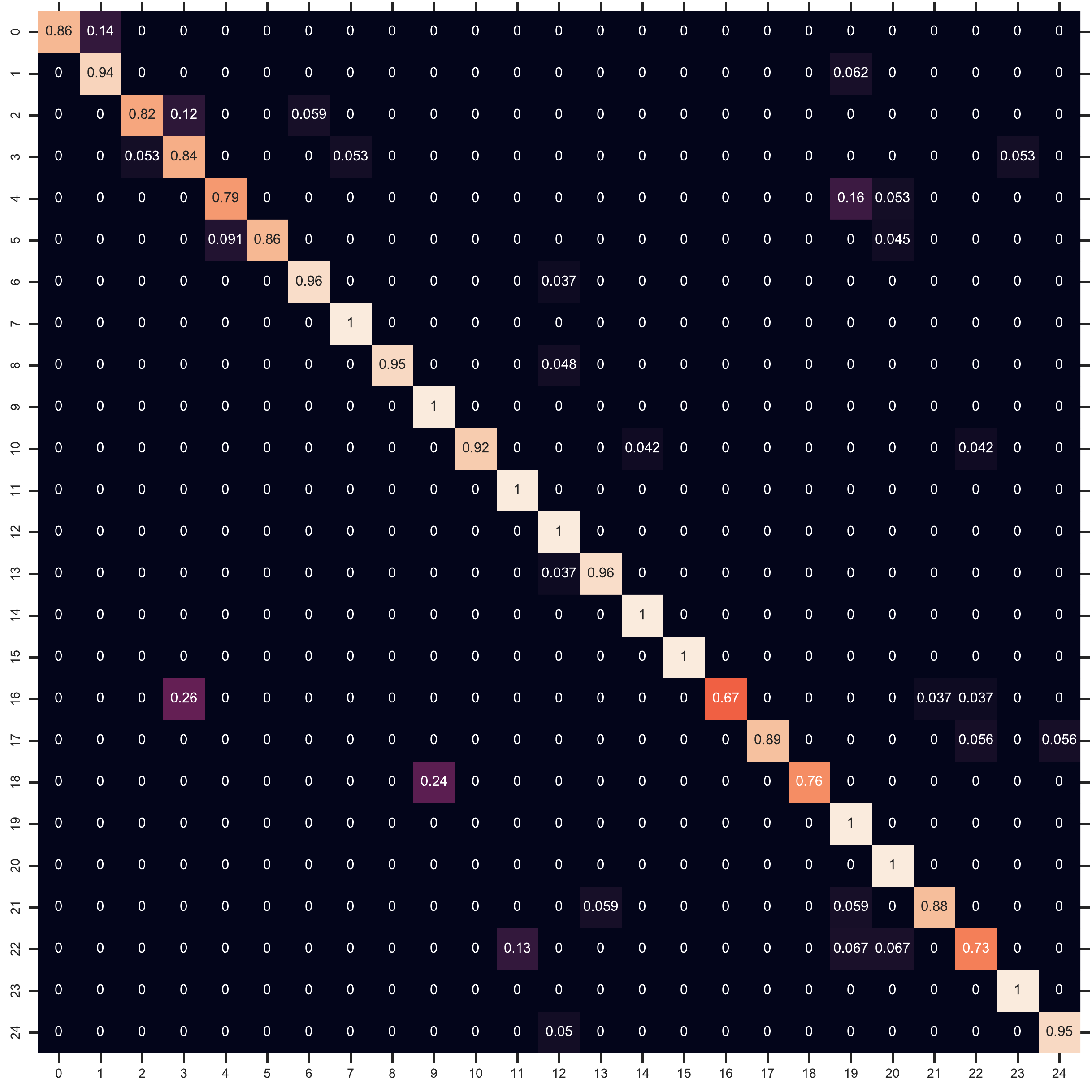} 
 \put (45,-5) {(a)}
\end{overpic}
\begin{overpic}[scale=0.15,tics=3]{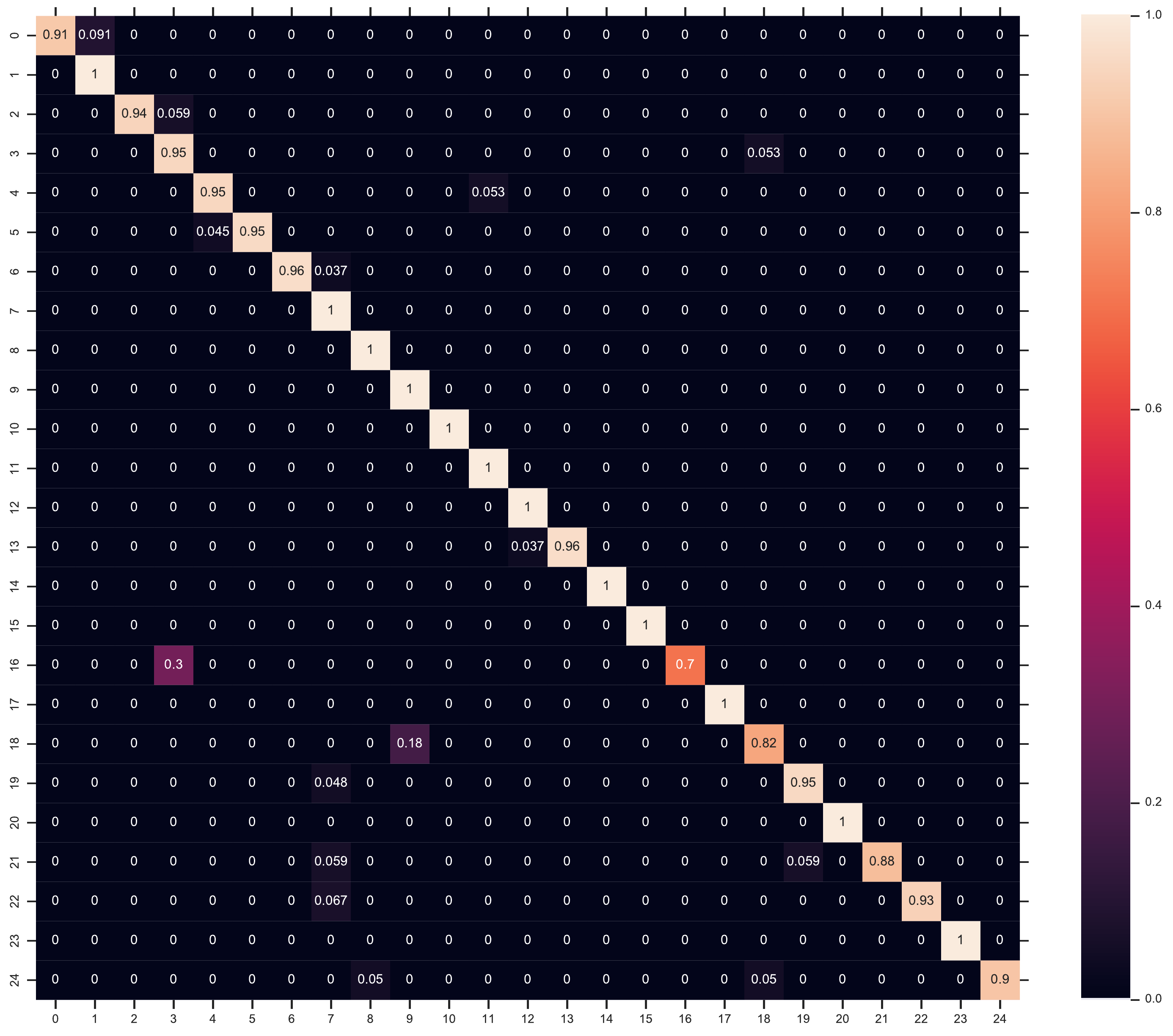} 
 \put (40,-5) {(b)}
\end{overpic}

\end{center}
\vspace{-1mm}
   \caption{The confusion matrices obtained by comparing the grand-truth labels and the predicted labels from the RGB network trained on the NVGesture dataset by (a) I3D~\cite{carreira2017quo} model, and  (b) our model. Best seen on the computer, in color and zoomed in.}
\label{fig:confusionmatrices}
\vspace{-2.5mm}
\end{figure}

\subsection{NVGesture Dataset}
In order to test our method on tasks with more than two modalities,  in this section, we report the classification results on the RGB, depth and optical flow modalities of the NVGesture dataset~\cite{molchanov2016online}.  The RGB and optical flow modalities are well-aligned in this dataset, however, the depth map includes a larger field of view (see Figure~\ref{fig:datasets} (b)).

Table~\ref{tbl:nv_unimodal} tabulates the results of our method in comparison with the recent state-of-the-art methods: HOG+HOG2, improved dense trajectories (iDT)~\cite{wang2016robust}, R3DCNN~\cite{molchanov2016online}, two-stream CNNs~\cite{simonyan2014two}, and C3D as well as human labeling accuracy.  The iDT~\cite{wang2016robust} method  is often recognized as the best performing hand-crafted method~\cite{tran2018closer}. However, we observe that similar to the previous experiments the 3D-CNN-based methods outperform the other hand gesture recognition methods, and among them, our method provides the top performance in all the modalities. 
This table confirms that our method can improve the unimodal test performance by leveraging the knowledge from multiple modalities in the training stage. This is despite the fact that the depth map in this dataset is not completely aligned with the RGB and optical flow maps.

Figure~\ref{fig:confusionmatrices} evaluates the coherence between the predicted labels and ground-truths in our method and compares it with I3D  for the RGB modality of the NVGesture dataset. This coherence is calculated by their confusion matrices. We observe that our method has less confusion between the input classes and provides generally a more diagonalized  confusion matrix.  This improvement is better observed in the first six classes.

 \begin{table}[t]
\begin{center}
%\resizebox{\linewidth}{!}{%
\begin{tabular}{ l c c c}
\hlineB{3}
%\cline{3-7}
& \multicolumn{3}{c}{\small{Testing modality}} \\
\cline{2-4} 
 Method   & RGB & Depth & Opt. Flow\\% & Fusion (RGB-D)\\%   & Avg.          $\pm$  std.      \\

\hline
HOG+HOG2~\cite{ohn2014hand}  &	24.5	& 	36.3	& 	-  \\%& - & - & -		 \\
Two Stream CNNs~\cite{simonyan2014two} &	54.6	& 	-	& 	68.0 \\%& - & - & -		 \\

C3D~\cite{tran2015learning} &	69.3	& 	78.8	& 	- \\%& - & - & -		 \\
iDT~\cite{wang2016robust} &	59.1	& 	-	& 	76.8 \\R3DCNN~\cite{molchanov2016online} &	74.1	& 	80.3	& 	77.8 \\

I3D~\cite{carreira2017quo} &	78.42	& 	82.28	& 	83.19 \\
MTUT~(ours)  &	\bf{81.33}	& 	\bf{84.85} & 	\bf{83.40} \\

\hline\hline
Human labeling accuracy: & \multicolumn{3}{c}{\small{88.4}} \\ 
\hlineB{3}
\end{tabular}
\vspace{.05mm}
\caption{Accuracies of different unimodal hand gesture methods on the NVGesture dataset~\cite{molchanov2016online}. The top performer is denoted by boldface.} \label{tbl:nv_unimodal}
\end{center}
\vspace{-5.5mm}
\end{table}

\subsection{Effect of Unimodal Improvements on Multimodal Fusion}
As previously discussed, our method is designed for embedding knowledge from multiple modalities in unimodal networks for improving their unimodal test performance.  In this section, we examine if the enhanced unimodal networks trained by our approach can also improve the accuracy of a decision level fusion that is calculated from the average of unimodal predictions.   The decision level fusion of different modality streams is currently the most common fusion technique in the top performer dynamic action recognition methods~\cite{carreira2017quo,tran2015learning,simonyan2014two}.   

In Table~\ref{tbl:multi_viva} and  Table~\ref{tbl:multi_ego} we compare the multimodal-fusion versions of our method (MTUT$^{\text{F}}$) to state-of-the-art multimodal hand gesture recognition systems tested on the VIVA hand gesture and EgoGesture datasets, respectively.   As can be seen, our method shows the top multimodal fusion performance on both datasets. These tables show that if multiple modalities are available at the test time, the improved performance of unimodal networks gained by training with our model can also result in an improved multimodal fusion performance in the test time. 

Similarity, in Table~\ref{tbl:multi_nv} we report the multimodal fusion results on the NVGesture dataset. Note that since this dataset includes three modalities, based on the modalities we include in the training stage, we report multiple versions of our method.  We report the version of our method that includes all three modalities in the training stage as MTUT$^{\text{F}}_{\text{all}}$, and the versions that only involve (RGB+Depth) and (RGB+Optical-Flow) in their training as MTUT$^{\text{F}}_{\text{RGB-D}}$ and MTUT$^{\text{F}}_{\text{RGB-OF}}$, respectively.    While all versions of our method outperform the other multimodal fusion methods in Table~\ref{tbl:multi_nv}, the performances of MTUT$^{\text{F}}_{\text{RGB-D}}$ and MTUT$^{\text{F}}_{\text{all}}$ in the fusion of RGB+Depth is worth highlighting.   MTUT$^{\text{F}}_{\text{all}}$ in this experiment has also been trained on the absent modality, the optical flow, while MTUT$^{\text{F}}_{\text{RGB-D}}$ has been only trained on the RGB and Depth modalities. We observe that MTUT$^{\text{F}}_{\text{all}}$ has successfully integrated the knowledge of the absent modality and provided a better performance at the test time.

\begin{table}[t]
\begin{center}
\resizebox{\linewidth}{!}{%
\begin{tabular}{ l c c}
\hlineB{3}
%\cline{3-7}
 Method   & Fused modalities & Accuracy\\% & Fusion (RGB-D)\\%   & Avg.          $\pm$  std.      \\

\hline
HOG+HOG2~\cite{ohn2014hand}  &	RGB+Depth & 	64.5		 \\
CNN:LRN~\cite{molchanov2015hand} &	RGB+Depth 	& 	74.4		 \\
CNN:LRN:HRN~\cite{molchanov2015hand} &	RGB+Depth 	& 	77.5		 \\
C3D~\cite{tran2015learning} &	RGB+Depth 	& 	77.4		 \\
I3D~\cite{carreira2017quo}       &	RGB+Depth & 	83.10 \\
MTUT$^{\text{F}}$~(ours) &	RGB+Depth & 	\bf{86.08}		 \\
\hlineB{3}
\end{tabular}}
\vspace{0.5mm}
\caption{Accuracies of different multimodal fusion-based  hand gesture methods on the VIVA dataset~\cite{ohn2014hand}. The top performer is denoted by boldface.} \label{tbl:multi_viva}
\end{center}
\vspace{-5mm}
\end{table}

\begin{table}[t]
\begin{center}
\resizebox{\linewidth}{!}{%
\begin{tabular}{ l c c}
\hlineB{3}
%\cline{3-7}
%\cline{2-3} 
 Method   & Fused modalities & Accuracy\\% & Fusion (RGB-D)\\%   & Avg.          $\pm$  std.      \\

\hline
VGG16~\cite{simonyan2014very} &	RGB+Depth & 	66.5		 \\
VGG16 + LSTM~\cite{donahue2015long} &	RGB+Depth	& 	81.4		 \\
C3D~\cite{tran2015learning} &	RGB+Depth & 	89.7		 \\
C3D+LSTM+RSTTM~\cite{cao2017egocentric} &	RGB+Depth & 	92.2		 \\
I3D~\cite{carreira2017quo} &	RGB+Depth 	& 	92.78		 \\
MTUT$^{\text{F}}$~(ours) &	RGB+Depth	& 	\bf{93.87}		 \\
\hlineB{3}
\end{tabular}}
\vspace{0.5mm}
\caption{Accuracies of different multimodal fusion hand gesture methods on the EgoGesture dataset~\cite{molchanov2016online}. The top performer is denoted by boldface.} \label{tbl:multi_ego}
\end{center}
\vspace{-5mm}
\end{table}

\begin{table}[t]
\begin{center}
\resizebox{\linewidth}{!}{%
\begin{tabular}{ l c c}
\hlineB{3}
%\cline{3-7}
 Method   & Fused modalities & Accuracy\\% & Fusion (RGB-D)\\%   & Avg.          $\pm$  std.      \\

\hline
HOG+HOG2 & RGB+Depth & 36.9  \\
I3D~\cite{carreira2017quo} &	RGB+Depth& 83.82  \\
MTUT$^{\text{F}}_{\text{RGB-D}}$~(ours) &	RGB+Depth	&  85.48  \\
MTUT$^{\text{F}}_{\text{all}}$~(ours) & RGB+Depth& \bf{86.10} \\
\hline
Two Stream CNNs~\cite{simonyan2014two} &	RGB+Opt. flow & 65.6  \\
iDT~\cite{wang2016robust} &	RGB+Opt. flow & 73.4 \\
I3D~\cite{carreira2017quo} &	RGB+Opt. flow& 84.43  \\
MTUT$^{\text{F}}_{\text{RGB-OF}}$~(ours) &	RGB+Opt. flow & \bf{85.48}\\
MTUT$^{\text{F}}_{\text{all}}$~(ours) &RGB+Opt. flow& \bf{85.48} 	 \\
\hline

R3DCNN~\cite{molchanov2016online} &	RGB+Depth+Opt. flow & 83.8		 \\
I3D~\cite{carreira2017quo} &	RGB+Depth+Opt. flow  &85.68		 \\
MTUT$^{\text{F}}_{\text{all}}$~(ours) & RGB+Depth+Opt. flow & \bf{86.93}	 \\
\hline\hline
Human labeling accuracy: & &88.4 \\ 
\hlineB{3}
\end{tabular}}
\vspace{0.5mm}
\caption{Accuracies of different multimodal fusion hand gesture methods on the NVGesture dataset~\cite{cao2017egocentric}. The top performer is denoted by boldface.} \label{tbl:multi_nv}
\end{center}
\vspace{-5.5mm}
\end{table}

\subsection{Analysis of the Network}
To understand the effects of some of our model choices, we explore the performance of some variations of our model on the VIVA dataset.  In particular, we compare our method with and without the \emph{focal regularization parameter} and the \emph{SSA} loss. Beside our I3D-based method, we analyze these variations on a different backbone network, C3D~\cite{tran2015learning} as well. C3D  is another recently proposed activity recognition architecture. We name this method MTUT$_{\text{C3D}}$.  Besides, we use C3D+SSA and I3D+SSA to refer to versions of our method with C3D and I3D backbones that contain a variation of the \emph{SSA} loss that does not have the \emph{focal regularization parameter}.    For MTUT$_{\text{C3D}}$ and C3D+SSA, we apply the SSA loss on feature maps of the last maxpooling layer (``$\texttt{MaxPool3d\_5}$'' ).

To provide a fair comparison setting, we train these networks from scratch on the VIVA dataset, and report their performances in Table~\ref{tbl:ablation}. As can be seen, the top performer is our I3D-based network with both SSA and focal regularization parameter.  Several interesting observations can be made from the results in Table~\ref{tbl:ablation}.  As the table reveals, the I3D-based methods generally perform better than the C3D-based methods. This coincides with the previous reports~\cite{carreira2017quo}.   In addition, C3D+SSA and I3D+SSA methods in the case of RGB networks show improvements and in the case of depth modality have comparable results as compared to their base networks C3D and I3D, respectively.  However, the top performers  in both modalities are the full version of our method applied on these base networks.    This clearly shows the importance of our focal regularization parameter in avoiding negative transfer when transferring the knowledge between the modalities.   Note that C3D, I3D and MTUT are trained from scratch in this experiment, while in the Table~\ref{tbl:viva_unimodal} we reported their performance on the networks trained with pre-trained weights.

 \begin{table}[t]
\begin{center}
%\resizebox{\linewidth}{!}{%
\begin{tabular}{ l c c}
\hlineB{3}
& \multicolumn{2}{c}{\small{Testing modality}} \\
\cline{2-3} 
 Method   & RGB & Depth\\%   & Avg.          $\pm$  std.      \\
\hline
 %\parbox[t]{2mm}{\multirow{6}{*}{\rotatebox[origin=c]{90}{C3D backbone}}}
   C3D             & 53.05  &  55.65  \\
   C3D+SSA             & 53.73  &  54.52   \\ 
  MTUT$_{\text{C3D}}$            & \textbf{56.56}  &  \textbf{58.71}   \\ 
  \hline 
   I3D             &  65.72	& 	67.30	  \\
   I3D+SSA             &  65.83  &  66.96   \\ 
   MTUT            &\bf{68.43}	& 	\bf{71.26}   \\ 
\hlineB{3}
\end{tabular}
\vspace{1mm}
\caption{ Comparison of variations of MTUT with C3D and I3D backbones trained from scratch.   } \label{tbl:ablation}
\end{center}
\vspace{-6.5mm}
\end{table}

\section{Conclusion}\label{sec:conclusion}
We presented a new framework to leverage the knowledge of multiple modalities when training unimodal networks that can independently work at the test time inference with improved accuracy.   Our model trains separate 3D-CNNs per available modalities, and shares their knowledge by the introduced \emph{spatiotemporal semantic alignment} loss.  We also regularized this loss with a \emph{focal regularization parameter} that ensures that only positive knowledge is transferred between the modality networks, and negative transfer is avoided.  Our experiments confirmed that our method can provide remarkable improvements to the unimodal networks at the test time.   We also showed that the enhanced unimodal networks that are trained with our method can contribute to an improved multimodal fusion performance at  test time as well.

The incorporation of our method for multimodal learning in other applications is a topic of further research.

%______________________________________THE END ____________

{\small
\bibliographystyle{ieee_fullname}
\bibliography{egbib}
}

\end{document}